%% file: main.tex
\documentclass[10pt,twocolumn,letterpaper]{article}

\usepackage{cvpr}              


\definecolor{cvprblue}{rgb}{0.21,0.49,0.74}
\usepackage[pagebackref,breaklinks,colorlinks,allcolors=cvprblue]{hyperref}

\def\paperID{00004} 
\def\confName{CVPR}
\def\confYear{2025}

\title{JOINT ATTITUDE ESTIMATION AND 3D NEURAL RECONSTRUCTION OF NON-COOPERATIVE SPACE OBJECTS}

\author{Clément Forray\\
CS Group\\
{\tt\small clement.forray@cs-soprasteria.com}
\and
Pauline Delporte\\
CS Group\\
{\tt\small pauline.delporte@cs-soprasteria.com}
\and
Nicolas Delaygue\\
CNES\\
{\tt\small nicolas.delaygue@cnes.fr}
\and 
Florence Genin\\
CNES\\
{\tt\small florence.genin@cnes.fr}
\and
Dawa Derksen\\
CNES\\
{\tt\small dawa.derksen@cnes.fr}
}

\begin{document}
\maketitle
\input{sec/0_abstract}    
\input{sec/1_intro}
\input{sec/2_related_work}
\input{sec/3_dataset}
\input{sec/4_nerf}
\input{sec/5_contributions}
\input{sec/6_implementation}
\input{sec/7_results}
\input{sec/8_conclusion}
\newpage
\small \bibliographystyle{ieeenat_fullname} \bibliography{main}

\end{document}

%% file: sec/0_abstract.tex
\begin{abstract}
Obtaining a better knowledge of the current state and behavior of objects orbiting Earth has proven to be essential for a range of applications such as active debris removal, in-orbit maintenance, or anomaly detection. 3D models represent a valuable source of information in the field of Space Situational Awareness (SSA). In this work, we leveraged Neural Radiance Fields (NeRF) to perform 3D reconstruction of non-cooperative space objects from simulated images. This scenario is challenging for NeRF models due to unusual camera characteristics and environmental conditions : monochromatic images, unknown object orientation, limited viewing angles, absence of diffuse lighting etc. In this work we focus primarly on the joint optimization of camera poses alongside the NeRF. Our experimental results show that the most accurate 3D reconstruction is achieved when training with successive images one-by-one. We estimate camera poses by optimizing an uniform rotation and use regularization to prevent successive poses from being too far apart.
\end{abstract}

%% file: sec/1_intro.tex
\section{Introduction}
\label{sec:intro}

The recent spur in space activities and the multiplication of satellites and debris orbiting Earth caused new challenges to emerge for the future. The European Space Agency (ESA) has been developing a Space Situational Awareness (SSA) program to address the technical, environmental and political issues at stake \cite{flohrer2017space}. This program includes a Space Surveillance and Tracking (SST) segment which aims at building the best knowledge possible of the state and position of every object orbiting our planet in the form of a catalog. These objects may include used rocket parts, collision debris as well as operational or off-service satellites. This data is crucial for numerous operational applications such as mission planning and analysis \cite{falco2021launch}, collision avoidance \cite{rongzhi2020spacecraft}, active debris removal \cite{liou2011active, palmerini2016guidelines}, in-orbit maintenance or anomaly detection \cite{tatsch2006orbit}. In this context, understanding the geometry and appearance of a space object helps to estimate its physical characteristics, movements and/or its capacities.

The emergence of agile satellite platforms and in-space observations \cite{zhang2024overview, olmos2013space} offers promising prospects compared to ground infrastructures and leads us to imagine new inspection scenarios. In this work, we studied an algorithm performing 3D reconstruction from a sequence of simulated in-orbit monochromatic optical images picturing a satellite moving in space. Our goal is to extract an exploitable 3D mesh or a point-cloud model for further processing and analysis.

As the target objects are non-cooperative, any prior information about the shape and movements is excluded from this work. Camera poses are thus unknown before training. We cannot resort to auxiliary sources of information such as masks or depth maps neither since these are not available in a general observation scenario.

Our approach relies on a custom training schedule strategy, we propose an incremental uniform attitude estimation method allowing joint geometry reconstruction and pose estimation for non-cooperative space objects. Our contribution is twofold : 

\begin{itemize}
    \item We adapted a NeRF model to the specificity of spaceborne mono-chromatic optical images implementing logarithmic tone mapping and opacity regularization losses.
    \item We developed a uniform rotation attitude estimation method by gradient descent with a custom training schedule strategy and a gradient-scaled multi-resolution positional encoding.
\end{itemize}

%% file: sec/2_related_work.tex
\section{Related Work}
\label{sec:related_work}

\subsection{3D Reconstruction in SSA}

The reflective properties of most objects orbiting Earth and the specificities of space lighting conditions make their appearance highly dependent on the viewing direction and solar angles. Space scenes considered in this paper represent high altitude scenarios and do not emulate Earth's albedo diffuse lighting. Consequently, the Sun is the only light source available. This is detrimental to the operation of classical structure-from-motion (SfM) algorithms \cite{hartley2003multiple, snavely2008modeling, snavely2006photo} which rely on point matching to retrieve the geometry of the target \cite{Papier1}.

Others have based their approach on Neural Radiance Fields (NeRF) \cite{Papier2} which have proven their efficiency and versatility in the field of 3D reconstruction in recent scientific literature \cite{zhu2023deep, yao2024neural}. The relevance of such models has been demonstrated in the context of SSA \cite{Papier3}. However, these works often rely on the scene orientation metadata \cite{Papier5}, feature-matching-based approximation \cite{Papier4, Papier6} or pre-trained pose-estimation networks \cite{Papier7}. The method introduced in this paper does not rely on the success of a prior pose-estimation model.

\subsection{Pose Estimation}

In the target object’s reference frame, the attitude estimation problem can be assimilated to a pose estimation problem for a set of cameras. Joint geometry reconstruction and pose estimation is particularly difficult in our case since space objects frequently show pseudo-symmetries which creates ambiguities in their possible orientation. Moreover, missing depth information can result in an inversion of the rotation direction as in Nobuyuki Kayahara's famous spinning dancer illusion \cite{troje2010viewing}.

Most pose-estimation NeRF models publications rely on auxiliary sources of data. Indeed, some recent papers make use of object masks or depth maps such as Nope-NeRF \cite{Papier8} and CT-NeRF \cite{Papier9} but we considered those were not available in our scenario. We thus focused on utilizing the sequential nature of the data and the proximity between consecutive views. We identified four main strategies.

The first approach is derived from Structure from Motion (SfM) algorithms and uses feature matching to estimate coarse camera poses before training a NeRF model \cite{Papier10, Papier11}. These solutions are highly dependent on the success of the preprocessing step, which often cannot be guaranteed in operational scenarios. The work of \cite{issitt2025optimal} showed SfM failed to converge in nearly 25\% of their tested datasets depending on the satellite type, the orbital configuration, the altitude and the distance from the camera. SfM methods are highly sensitive to symmetries and the lack of distinctive features but also to the low parallax induced by coplanar orbits and cameras with limited angles of view (AOV).

The second approach relies on pre-trained models such as pose estimation networks to replace the SfM pipeline and initialize the camera poses \cite{Papier7, Papier12}. It seems difficult to build a representative dataset to train such a model considering the low quantity of available real-world data, and the diversity of appearance and shape of space objects. Some cases may cause out-of-distribution errors. 

The third approach uses Generative Radiance Fields (GRAF) \cite{schwarz2020graf} to perform the camera poses estimation by training a NeRF in an adversarial fashion \cite{Papier3}. However, adversarial training deteriorates the overall geometry reconstruction quality despite a large diversity of training views and can even result in mode collapse.

Finally, a few methods were developed with custom training strategies using only the sequential nature of our data and optimizing camera poses from scratch jointly with a NeRF model. LU-NeRF \cite{Papier13} consists in training local NeRF models for small groups of consecutive views and then merging them together. CF-NeRF \cite{Papier14} uses an incremental training strategy, introducing camera views sequentially during the optimization of the NeRF model. Both models assume consecutive cameras poses can not be too far apart which allows a restriction of the search space. Similarly to CF-NeRF, we introduce a simple incremental training strategy so that new camera views are made available progressively while optimizing the NeRF model. Each new camera pose can thus be initialized close to its expected position.

%% file: sec/3_dataset.tex
\section{Dataset}
\label{sec:dataset}

Our method was developed using a sequence of optical images generated with simulation tools provided by the French National Center of Space Studies (CNES). We emulated data collected by an agile inspection satellite carrying a panchromatic monocular camera located 2 kilometers away from the target object with centimetric resolution. We consider the absolute position and orientation of the observer as well as the camera intrinsics to be known during the whole duration of the sequence. However, the attitude maneuvers of the target remain inaccessible since we need to deal with non-cooperative objects as well.

Thus, we established a few hypotheses and defined three scenarios with increasing levels of difficulty presented in section \ref{subsec:xpsetup}. We used a noiseless background, empty and free of stars in all images. Neither the Earth, the Sun or any other space body appear in the camera field except the observed object. Images do not show saturation or glare artefacts. 

\begin{figure}[h]
  \centering
  \includegraphics[width=0.9\linewidth]{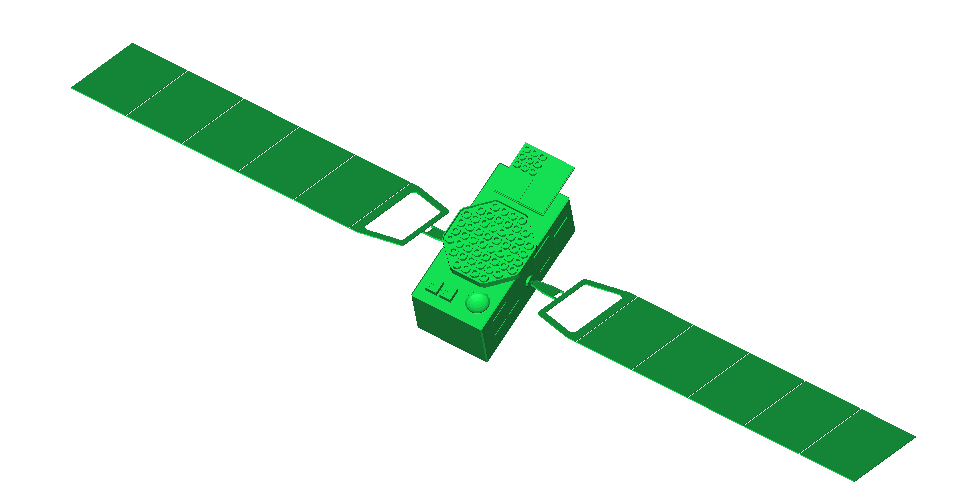}
  \caption{Reference Galileo mesh (CloudCompare \cite{cloudcompare})}
  \label{fig:galileo_mesh}
\end{figure}

The sequence used in this work depicts a rigid satellite describing an uniform rotation motion observed by a fixed camera located between the satellite and the Sun. This simple yet realistic configuration makes the lighting conditions simpler for the model and limits the amount of surface in the shadows. We chose the Galileo satellite mesh model in \cref{fig:galileo_mesh} as the target object for our experiments.

The dataset is composed of a sequence of 109 mono-chromatic images of size 1024x1024 representing the satellite performing slightly more than a full rotation. 99 are used for training and 10 are kept for validation, equally distributed among the available viewing angles. 

%% file: sec/4_nerf.tex
\section{Neural Radiance Fields}
\label{sec:nerf}

NeRF models belong to the family of implicit neural representations. Their general purpose is to learn a 3-dimensional scene as a radiance field implicitly encoded by a neural network. 3D reconstruction is performed by training a fully connected network (MLP) to estimate the density \(\sigma\) and radiance c of every point \((x, y, z)\) in the 3D scene from any angle \((\theta, \phi)\). A complete training is thus required for every new scene. 

\begin{figure}[h]
  \centering
  \includegraphics[width=0.9\linewidth]{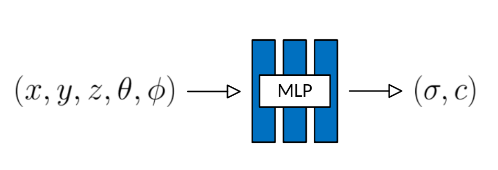}
   \label{fig:nerf}
   \caption{NeRF model representation}
\end{figure}

This representation allows to generate novel views of the scene from any arbitrary angle through volumetric rendering. The value of a pixel in the produced image corresponds to the sum of the radiances of the points along the line of sight weighted by their visibility as shown in \cref{eq:3}. The visibility of a given point being determined by its own density and that of the points separating it from the camera in accordance with \cref{eq:1} and \cref{eq:2}. Similarly, it is possible to generate a depth map of the scene by summing the distances from the camera weighted with the visibilities of the sampled points, \cref{eq:4}. 

We define the opacity \(\alpha\) as a function of the estimated density \(\sigma\) and the distance from the previous point sample \(\delta\) along the ray: 

\begin{equation}
\label{eq:1}
    \alpha_i = 1 - exp(- \sigma_i \delta_i) 
\end{equation}

The visibility weight w given to a point sample can be written as: 

\begin{equation}
\label{eq:2}
    w_i = \alpha_i  \prod_{k < i} \left ( 1 - \alpha_k \right ) 
\end{equation}

The rendering of the radiance \(C\) and the depth \(D\) estimated for a ray \(r\) is therefore: 

\begin{equation}
\label{eq:3}
    C(r) = \sum_{x_i \in r} w_i c_i 
\end{equation}

\begin{equation}
\label{eq:4}
    D(r) = \sum_{x_i \in r} w_i x_i 
\end{equation}

As a combination of simple differentiable operations, the rendering process is itself differentiable. Therefore, NeRF models can be trained in a self-supervised fashion with a simple L2 norm between the images of the dataset and the rendered predictions of the model for the same angles. Rendered depth maps can be used to extract a point-cloud representation of the learned scene geometry which is an exploitable format for further analysis and operational applications. Furthermore, a mesh model can also be produced from the NeRF scene representation using the marching cubes algorithm.

%% file: sec/5_contributions.tex
\section{Contributions}
\label{sec:contributions}

\subsection{Model Architecture}

\begin{figure*}[t]
  \centering
  \includegraphics[width=0.85\linewidth]{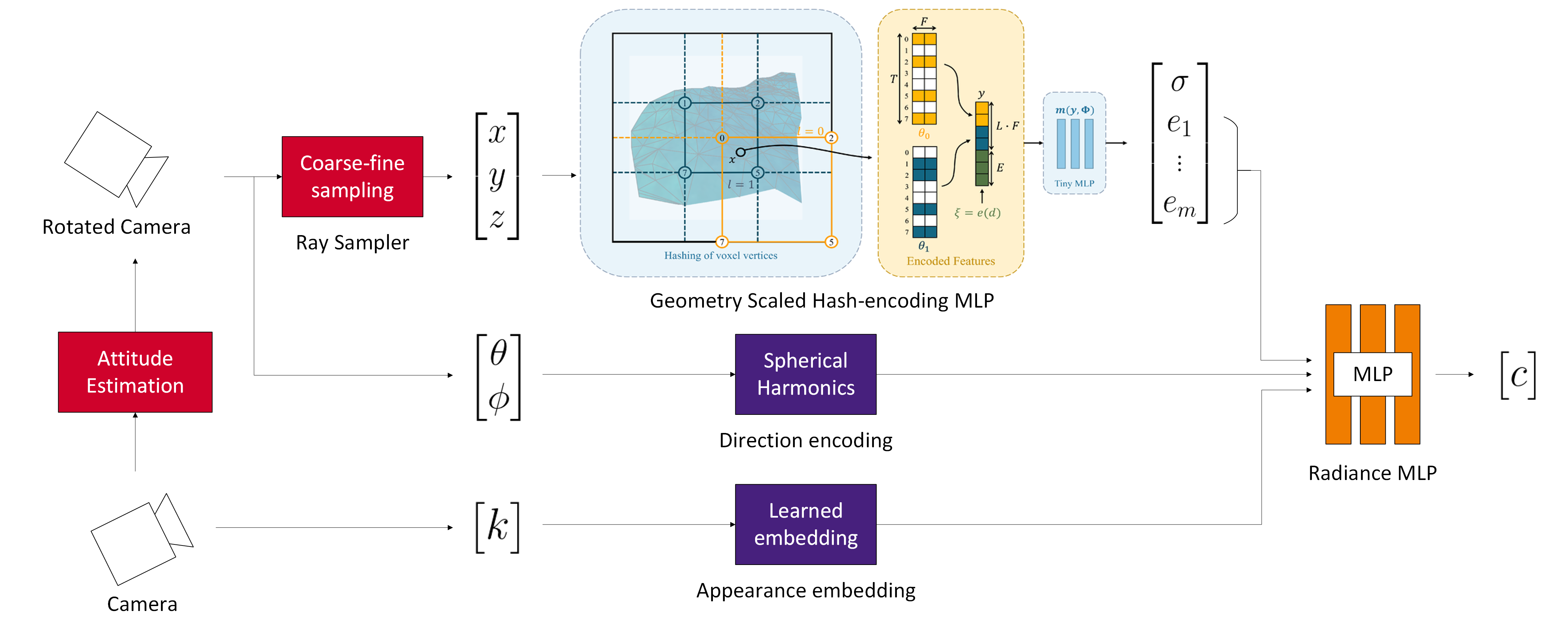}
   \label{fig:model_archi_base}
   \caption{Architecture of the model (hash-encoding source \cite{Papier16})}
\end{figure*}

At each iteration, we start by sampling a batch of pixels from the dataset. 3D points are sampled along the lines of sight of each pixel using a simple coarse-fine sampling strategy introduced in \cite{Papier2}. A first set of coarse samples is selected by randomly jittering uniformly distributed points along the ray. A set of fine samples is then chosen using the density distribution predicted from the first set. The fine samples are located closer to the scene surfaces and help the model focus on important areas, increasing the level of details while maintaining a reasonable number of samples. We chose this simple sampling strategy to ensure compatibility with our attitude estimation module.

The samples are used as input to a multiscale hash-encoding following the work of instant-NGP \cite{Papier15}, significantly accelerating the inference process. A small MLP predicts an opacity value as well as a geometry embedding. A second MLP predicts mono-chromatic radiance using this embedding along with an additional direction encoding which decomposes viewing angles with spherical harmonics. Indeed, since space objects often include surfaces with specular behavior, the radiance of the object may vary according to the observation and lighting angles. In our scenario, the observation and lighting angles are entirely correlated since the Sun remains behind the camera at all times. Therefore, radiance can be modeled as a function of the viewing angles only.

A learned appearance embedding was also included in the radiance model to make it robust to minor inconsistencies that may exist between views under similar angles by using the image index in the dataset. We observed removing it was detrimental to the visual quality of the model, probably due to shadows areas resulting from slight misalignment between the satellite, the camera and the Sun.

\subsection{Attitude Estimation}

We defined the scene’s reference frame as the orthonormal frame attached to the target object whose axes align with the orientation of the camera in the first view of the sequence. In this reference frame, the trajectory of the camera can be represented by a 3D rotation around an axis passing through the origin. Everything works as if the camera was in motion around a static scene which is the default configuration for NeRF model training. Since we hypothesized the camera and the Sun are fixed in the absolute reference frame, the lighting conditions are perfectly correlated to the viewing angle. In our case, the Sun will always be located directly behind the camera which ensures optimal lighting conditions for the reconstruction. 

Our work is based on Barf (Bundle Adjusting neural Radiance Fields) \cite{Papier17} and its adaptation to hash-encodings \cite{Papier18} and consists in using a coarse-to-fine camera registration strategy. A 3D rotation is optimized for each image in the dataset, jointly trained with the NeRF model using gradient descent. Rotations are parametrized according to the axis-angle representation as a 3D vector whose direction is the rotation’s axis and its norm represents the angle. The proposed scaled hash encoding is implemented so that the gradients of the high frequency features are masked at the beginning of the training according to a user-defined schedule. This way, the model is compelled to gradually increase its level of details during the optimization process and the attitude estimation optimizer only receives low resolution spatial gradients at first. Gradients for pose optimization are thus smoother at the beginning of the training when the optimizer steps are the largest which showed crucial for model convergence.  

However, this method works only if the camera poses are initialized not too far from their true positions. We empirically find the limit to be 15° approximately. We exploit the sequential nature of the data to define an initialization stage in which the model is trained with the first few images of the dataset only since the first camera poses are expected to be quite close to each other. Gradually, the remaining views are fed to the model during an incremental training stage. 

Under the hypothesis of uniform rotation and the knowledge of acquisition times for all the images, it is possible to parameterize a unique rotation for all the cameras with only 3 parameters. This strategy makes the training more stable.

\subsection{Logarithmic tone-mapping}

Due to the specular nature of the object surfaces, large variations in radiance can appear between pixels in our dataset. This is detrimental to the convergence of a NeRF model as errors committed on the lightest pixels will be over-penalized and outweigh those of lower pixel values. We thus introduced a logarithmic tone-mapping which is applied on all images of our dataset before training. 

\begin{equation}
    x  \mapsto \frac{log\left ( 1 + x \right )}{log\left ( 1 + M \right )} 
\end{equation}

With \(M\) greater than the maximum radiance value in the dataset.

\subsection{Regularization}

We observed our model often converges toward transparent solutions. Indeed, with mono-chromatic images and an empty black background, there is no radiometric difference between a pixel showing a bright transparent surface and a darker opaque one. We hypothesized transparent materials were quite rare amongst space objects and thus introduced several regularization loss terms to guide our model toward more plausible solutions. 

The opacity loss encourages weights accumulation on a camera ray to be either 0 or 1: 

\begin{equation}
    L_{opacity}(r) = - \left ( \sum_{x_i \in r} w_i  \right ). log\left ( \sum_{x_i \in r} w_i  \right ) 
\end{equation}

Additionally, penalizing bright surfaces also encourages surfaces to be less transparent. Hence the radiance regularization loss term: 

\begin{equation}
    L_{radiance}(r) = \sum_{x_i \in r} c_i  
\end{equation}

Besides, training the model jointly with camera pose estimation requires additional regularization. We use the distortion loss introduced in \cite{barron2022mip} and a simple L1 regularization on the camera rotation parameters. 

\subsection{Metrics}
\label{subsec:metrics}

We assessed the quality of our method and visualized errors by putting aside a dozen images from the dataset during training. Peak Signal to Noise Ratio (PSNR) and Learned Perceptual Image Patch Similarity (LPIPS) are common metrics in the field of computer vision and appeared relevant for the evaluation of our model for novel-view synthesis. However, we found that the Structural Similarity Index Measure (SSIM) was not fit to evaluate our data and did not reflect the visual quality of the rendered images. We explain this behavior by the noiseless background in our simulated data, tricking the SSIM to find a black image has a better score than our model’s predictions. The SSIM indeed compares the local mean, variance and covariance of two images and is therefore sensitive to noiseless data.  

Moreover, we introduce 3D metrics to quantify absolute geometry error. We exported a point-cloud representation of the scene using rendered depth maps from every angle available during training and filtered outliers. Statistical outlier removal is performed using a threshold on the distance of a point to its K nearest neighbors compared to the average distance within the point cloud. We defined two error metrics between the obtained point cloud and the reference point cloud. We compute the average distance between each predicted point to the nearest reference point and the other way around. The combination of those two distances reflects 3D geometry error in the form of precision and recall scores. Indeed, the distance from the prediction to the reference is similar to a precision score because it reflects how close predicted points are from the truth. On the other hand, the distance from the reference to the prediction can be compared to a recall score as it reveals possible missed parts compared to the true geometry.



\begin{equation}
    Precision = \frac{1}{N_{pred}} \sum_{x_{pred}} \min_{x_{ref}} \left ( \left\| x_{pred} - x_{ref} \right\|_2 \right )
\end{equation}

\begin{equation}
    Recall = \frac{1}{N_{ref}} \sum_{x_{ref}} \min_{x_{pred}} \left ( \left\| x_{pred} - x_{ref} \right\|_2 \right )
\end{equation}

%% file: sec/6_implementation.tex
\section{Implementation}
\label{sec:implementation}

We chose the open-source framework Nerfstudio \cite{nerfstudio} for the implementation of our solution. This framework is built on torch and Nvidia tiny-cuda-nn and provides flexible layers of abstraction for the development of volumetric and surface models while benefiting from cuda optimized acceleration. 

The derivation of our 3D metrics was performed using CloudCompare \cite{cloudcompare} software and the point-cloud distance tool. The graphical interface helped us visualize and superimpose our point-clouds. 

All training runs were performed thanks to the CNES HPC cluster, using a NVIDIA Tesla V100 32G GPU.  

%% file: sec/7_results.tex
\section{Results}
\label{sec:results}

\subsection{Experimental setup}
\label{subsec:xpsetup}

We derived 3 experiments to test and validate our approach. For each of them, we trained our model for 30000 steps. The coarse-to-fine gradient-scaled hash encoding is configured with a cosine schedule so that the model is allowed to learn higher frequencies progressively during the first 10000 steps. We observed this schedule should be tuned in adequacy with the learning rate of the attitude estimation optimizer so that larger steps are made when the gradients are smoother. Otherwise, the gradients guiding the attitude estimator will rapidly contain noise due to high spatial frequency details that will eventually cause the training to collapse.

In the first experiment (\textit{baseline}), we trained the model using the true camera positions to serve as a reference. Our synthetic dataset provides the orientation of the observed object during the simulated sequence. We can thus initialize the camera poses in the object’s reference frame and train our model without activating the attitude estimation module. These settings were designed to serve as a baseline for the best reconstruction quality reachable with our model architecture. 

In the second experiment (\textit{indep}), we added small random rotations to the initial camera poses to evaluate the ability of our method to compensate errors in prior pose estimation. The rotation axis of the perturbations are uniformly distributed and the angles follow a gaussian distribution with a mean of 8° and a standard deviation of 2°. We estimate the camera poses by optimizing independent 3D rotations jointly with the NeRF model during training. This experiment shows the ability of our model to retrieve camera poses when they are initialized not too far away from the truth. 

Finally, we put our method to the test without any priors on the initial camera poses (\textit{global}). First, we train our model using only the first 8 views from the image sequence during a 10000 steps initialization stage, then we introduce a new image every 100 steps during a 20000 steps incremental stage until the end of the training. This time, a single rotation is estimated for all cameras and poses are derived using the acquisition time so that the resulting movement is a uniform circular motion. The initial value for the estimated rotation is zero at the beginning of the training. 

\subsection{Novel view synthesis}

\begin{figure*}[t]
  \centering
  \includegraphics[width=0.9\linewidth]{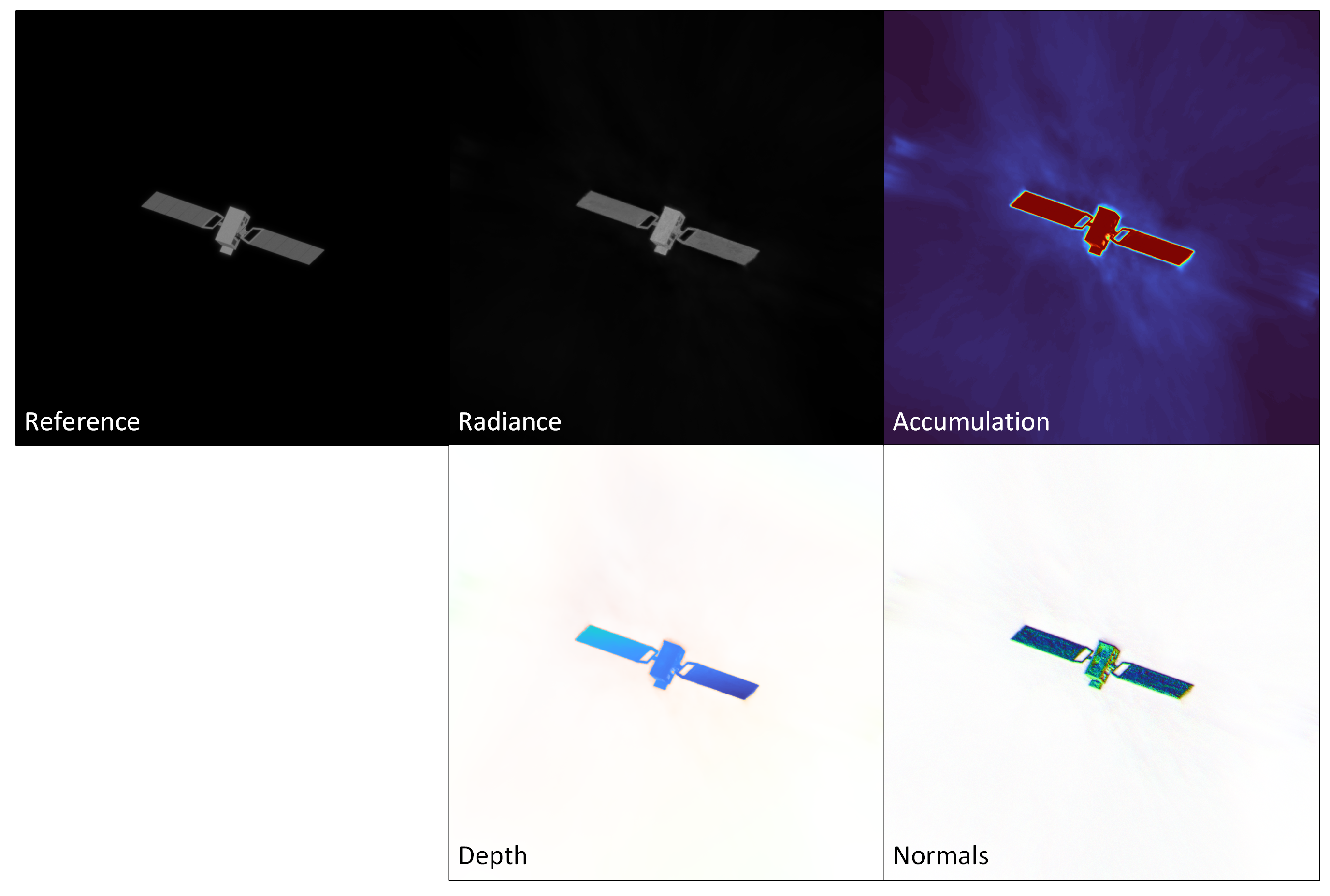}
   \caption{Validation view (reference image compared to rendered radiance, accumulation, depth map and surface normal directions)}
   \label{fig:inference_images}
\end{figure*}

All three experiments show comparable results according to our radiometric metrics (\cref{tab:2D_metrics}). It suggests our attitude estimation strategy works well and does not cause much quality loss compared to the baseline. Rendering results for an evaluation image in the global attitude estimation configuration are reported in \cref{fig:inference_images}. The figure depicts the reference image, the rendered radiance, accumulation, depth map and the surface normals orientations. Accumulation is the sum of the samples weights and represents the opacity of the ray. Empty space is displayed in deep blue whereas opaque surfaces are shown in red. The surface normals are derived from the density field with implicit differentiation and the color in the figure represents their directions. 

The model reconstruction seems to achieve high visual quality with rendered radiance, the object’s symmetry is respected, and several details are represented on the side of the satellite. The rendered accumulation shows shows that the opacity of the object is properly learned. However, defects can be noticed around the satellite as the surrounding void  does not appear totally transparent. Rendered depth and normals maps also appear correct. 

\begin{table}[h]
  \centering
  \begin{tabular}{c|c|c|}
    & \textbf{LPIPS \( \downarrow \)} & \textbf{PNSR \( \uparrow \)} \\
    \hline
    Baseline & 0.0123 ± 0.0025 & \underline{34.27} ± \underline{1.013} \\
    \hline
    Indep & \underline{0.0114} ± \underline{0.0038} & 33.88 ± 2.802 \\
    \hline
    Global & 0.0158 ± 0.0027 & 33.12 ± 0.641 \\
    \bottomrule
  \end{tabular}
  \caption{Radiometric error metrics}
  \label{tab:2D_metrics}
\end{table}

\subsection{3D reconstruction}
\label{subsec:3dreconstruction}

We measured errors in geometry with the 3D metrics defined in \cref{subsec:metrics}. As opposed to the radiometric results, the global optimization mode significantly outperforms the independent optimization mode with a precision of \textbf{17.2 mm} on average with a standard deviation of \textbf{12.5 mm} against \textbf{26.2 mm} with \textbf{51.7 mm} of standard deviation. Both recall metrics remain close to the baseline.

\begin{table}[h]
  \centering
  \begin{tabular}{c|c|c|c|c|}
    & \textbf{Precision (mm) \( \downarrow \)}& \textbf{Recall (mm) \( \downarrow \)} \\
    \hline
    Baseline & \underline{12.3} ± \underline{18.9} & \underline{5.18} ± \underline{4.30} \\
    \hline
    Indep & 26.2 ± 51.7 & 6.49 ± 5.64 \\
    \hline
    Global & 17.2 ± 12.5 & 8.65 ± 7.65 \\
    \bottomrule
  \end{tabular}
  \caption{Geometry error 3D metrics}
  \label{tab:3D metrics}
\end{table}

\begin{figure}[h]
  \centering
  \includegraphics[width=1 \linewidth]{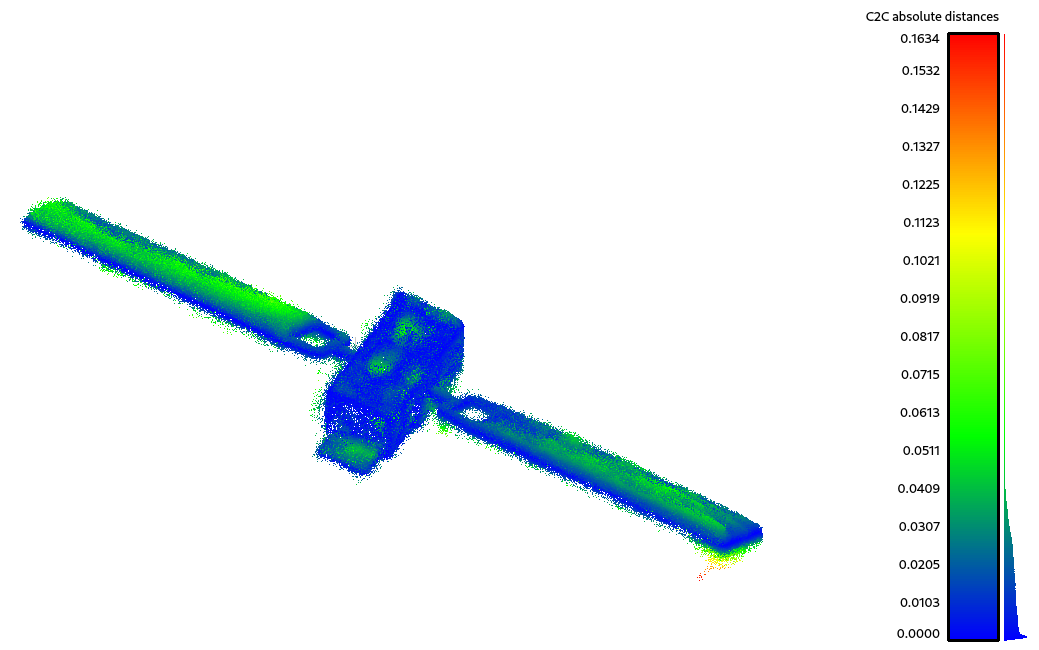}
  \caption{Precision point cloud  (in meters)}
  \label{fig:Precision point cloud (in meters)}
\end{figure}

\begin{figure}[h]
  \centering
  \includegraphics[width=1 \linewidth]{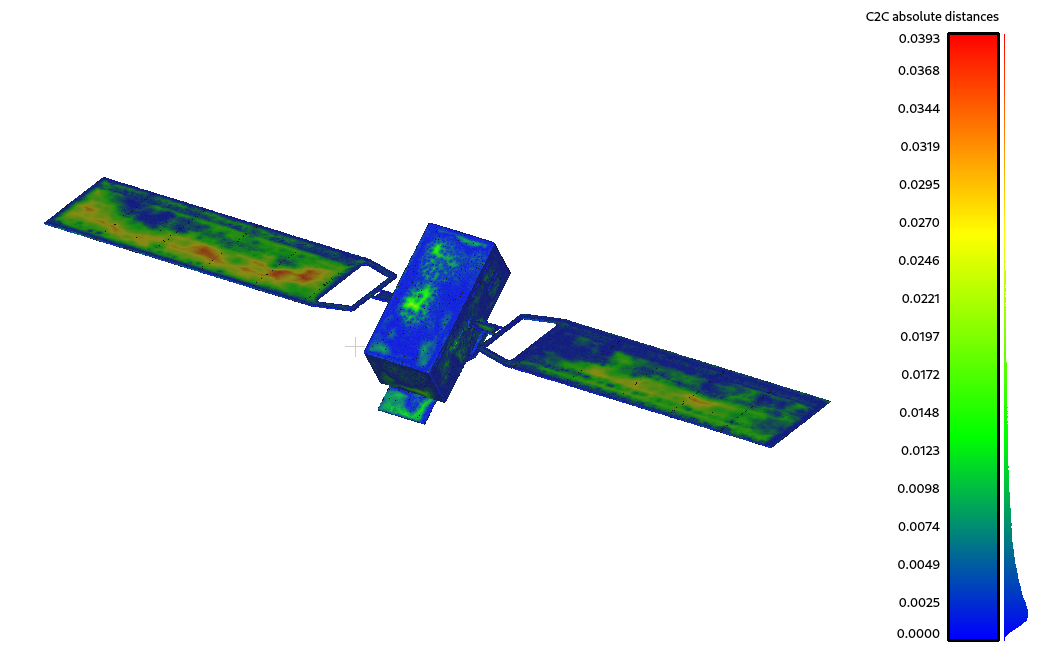}
  \caption{Recall point cloud (in meters)}
  \label{fig:Recall}
\end{figure}

\label{subsec:novelview}
\subsection{Pose estimation}
\label{subsec:posestim}

We compare the estimated camera poses to the poses metadata of our dataset. \Cref{tab:pose_estimation_error} reports the angular error in degrees measured at the end of the training. We show that \textit{global} mode is more robust and provides better median and maximal angular errors than the \textit{indep} mode. We also included the angular error measured for a SfM pipeline, proving the interest of our method in this challenging scenario.

\begin{table}[h]
  \centering
  \begin{tabular}{c|c|c|}
    \textbf{Angular error} & \textbf{Maximal \( \downarrow \)} & \textbf{Median \( \downarrow \)} \\
    \hline
    SfM & 174° & 83° \\
    \hline
    Indep & 15° & 5° \\
    \hline
    Global & \underline{3°} & \underline{1.6°} \\
    \bottomrule
  \end{tabular}
  \caption{Pose estimation error}
  \label{tab:pose_estimation_error}
\end{table}

%% file: sec/8_conclusion.tex
\section{Conclusion}
\label{sec:conclusion}

Our method allows joint attitude estimation and 3D reconstruction of non-cooperative space objects from a sequence of mono-chromatic optical images captured by an agile observation satellite. It was designed for a simple yet realistic scenario where the attitude of the target follows a uniform rotation motion and the camera remains stationary to provide optimal lighting conditions. A point cloud and a mesh representation of the geometry can be produced from the learned 3D model which allows further processing in SSA applications. 

We validated the approach using a simulated dataset. We used quantitative metrics to measure errors in terms of novel views synthesis, pose estimation and produced point cloud. 

Our method leverages Neural Radiance Fields (NeRF) models to learn the geometry of the target object. We adapted our approach to SSA conditions by implementing an attitude estimation module which is optimized jointly with the NeRF model during training. The model doesn’t rely on any auxiliary data sources and only assumes a uniform rotation motion. When this hypothesis is not verified, our experiments demonstrate that the model could work with coarse camera poses. 

\section{Limitations}
\label{sec:limitation}

As mentioned above, our method is limited to uniform rotation attitude motions and ideal lighting conditions. We believe our approach can be adapted to cover broader operational conditions.  

Besides, our model could benefit from improvements in robustness, as we found the results to be somewhat dependent on the choice of hyperparameters. The tuning of the regularization weights, the scaled hash-encoding schedule and the learning rate of the attitude estimator were the most critical.